\pdfoutput=1
\documentclass[lettersize,journal]{IEEEtran}
\usepackage{amsmath,amsfonts}
\usepackage[ruled]{algorithm2e}
\usepackage{algpseudocode}
\usepackage{booktabs}
\usepackage{siunitx}
\usepackage[table]{xcolor}
\usepackage{array}
\usepackage{textcomp}
\usepackage{stfloats}
\usepackage{url}
\usepackage{verbatim}
\usepackage{graphicx}
\usepackage{cite}
\usepackage{csquotes}
\usepackage{enumitem}
\newcommand*{\ditto}{---\texttt{"}---}
\usepackage{hyperref}
\usepackage{pifont}
\usepackage[table]{xcolor}
\begin{document}

\title{Efficient CNNs via Passive Filter Pruning}

\author{Arshdeep Singh,~\IEEEmembership{Member,~IEEE,},  Mark D Plumbley, ~\IEEEmembership{Fellow,~IEEE,}
\thanks{Arshdeep Singh and Mark D. Plumbley are with the Centre for Vision, Speech and Signal Processing, University of Surrey, Guildford GU2 7XH, U.K. (e-mail: arshdeep.singh@surrey.ac.uk; m.plumbley@surrey.ac.uk).}

}



\maketitle

\begin{abstract}

Convolutional neural networks (CNNs) have shown state-of-the-art performance in various applications. However, CNNs are resource-hungry due to their requirement of high computational complexity and memory storage. Recent efforts toward achieving  computational efficiency in CNNs involve filter pruning methods that eliminate some of the filters in CNNs based on the \enquote{importance} of the filters. The majority of  existing filter pruning methods are either \enquote{active}, which use a dataset and generate feature maps to quantify filter importance, or \enquote{passive}, which compute filter importance  using entry-wise norm of the filters without involving data.
Under a high pruning ratio where  large number of filters are to be pruned from the network, the entry-wise norm methods eliminate relatively smaller norm filters without considering  the significance of the filters in producing the node output,
resulting in degradation in the performance. To address this, we present a passive filter pruning method where the filters are pruned based on their contribution in producing output by considering the operator norm of the filters. 
The proposed pruning method generalizes better across various CNNs compared to that of the entry-wise norm-based pruning methods. In comparison to the existing active filter pruning methods, the proposed pruning method is at least 4.5 times faster in computing filter importance and is able to achieve similar performance compared to that of the active filter pruning methods. The efficacy of the proposed pruning method is evaluated on audio scene classification 
and image classification
using various CNNs architecture such as VGGish, DCASE21\_Net, VGG-16 and ResNet-50.

\end{abstract}


\section{Introduction}
\IEEEPARstart{C}{onvolutional} neural networks (CNNs) have shown great success and exhibit state-of-the-art performance when compared to traditional hand-crafted methods in many domains \cite{gu2018recent} including computer vision \cite{kyrkou2019deep,wang2021comparative} and audio classification \cite{kong2020panns}. While CNNs are highly effective in solving non-linear complex tasks \cite{denton2014exploiting},  it may be challenging to deploy large-scale CNNs on resource-constrained devices such as mobile phones or internet of things (IoT) devices, owing to high computational costs during inference and the memory requirement for CNNs \cite{simonyan2014very, krizhevsky2012imagenet}. Thus, the issue of reducing the size  and the computational cost of CNNs has drawn a significant amount of attention in the research community. 

Recent efforts toward reducing the computational complexity of CNNs include pruning methods, where a set of parameters, such as     weights or convolutional filters, are eliminated from the CNNs. These pruning methods are motivated by the existence of redundant parameters  \cite{denil2013predicting, livni2014computational} in CNNs that only yield extra computations without contributing much in performance  \cite{frankle2018lottery}.
For example, Li et al. \cite{li2016pruning} found that 64\% of the parameters, contributing approximately 34\% of the computation time, are redundant. Eliminating such redundant parameters from CNNs provides small CNNs that perform similarly to the original CNNs while reducing the number of  computations and the memory requirement compared to the original CNNs.

While eliminating individual weights from an unpruned CNN may result in a  network with fewer parameters, the resulting network may be unstructured and may not be straightforward to run more efficiently. Practical acceleration of such unstructured sparse pruned networks is limited due to the random connections \cite{luo2018thinet}. Moreover, the unstructured sparse networks are not supported by off-the-shelf libraries and require specialised software or hardware for speed-up \cite{wen2016learning,han2016eie}. To address this unstructured pruning problem,  several filter pruning methods have been proposed which eliminates  convolutional filters at once, resulting in a structured pruned network  that does not require additional resources for speed-up. 


In these structured filter pruning methods, the \enquote{importance} of a convolutional filter, used to decide if the filter retains most performance, is measured using either active or passive methods. Active filter pruning methods  \cite{luo2017entropy,lin2020hrank,liu2017learning,lin2019towards} involve a dataset to compute the importance of filters. On the other hand, passive filter pruning methods \cite{li2016pruning,he2019filter} are data-free and only use the parameters of the filters without involving any dataset to compute the importance of the filters, and hence are relatively simpler than active filter pruning methods. More details on active and passive filter pruning methods is explained in Section \ref{sec: filter pruning methods}. Typically, existing passive filter pruning methods compute absolute sum of coefficients in the convolutional filter  to obtain importance of the filter without considering how significantly the convolutional filter produces output, resulting in degradation in performance particularly when there are large number of filters to be pruned from the network, due to retaining the filters producing less significant output. More detail on problems with existing passive filter pruning methods is explained in Section \ref{sec: problems with passive filter pruning}.
In this paper, we hypothesize that while computing importance of convolutional filters in CNNs, it is useful to incorporate the geometric relationship on how an input gets transformed to output by the convolutional filters for considering the significance of convolutional filters in producing output as convolutional filters producing low significance output contains less information, and thus can be eliminated.

The contribution of the paper includes, (1)  we propose a novel passive filter pruning method that considers  implicitly the contribution of the convolutional filters in producing output. (2) We validate the proposed pruning framework across  audio and image domains using various CNN architectures including DCASE21\_Net, VGGish, VGG-16 and ResNet-50.
(3) We show that our pruning method significantly reduces  the computational complexity of CNNs in terms of the number of parameters and the multiply-accumulate operations (MACs) significantly and provides similar accuracy compared to the unpruned CNNs. (4) We show that considering contribution of the  convolutional filters in producing output while computing importance of filters result in better generalisation across audio and image domains compared to that of the considering absolute sum of  convolutional filters coefficients as importance scores particularly when a large number of filters are pruned. In comparison to methods involving dataset in pruning, the proposed pruning method is at least 4.5 times faster and requires no extra memory overhead to generate outputs from convolutional filters, and is able to achieve competitive performance to that of pruning methods involving dataset in pruning. (5) We have released the source code\footnote{Link to codes: \url{https://github.com/Arshdeep-Singh-Boparai/Efficient_CNNs_passive_filter_pruning.git}} for operator norm based pruning method and the pruned networks obtained from the unpruned network.

This paper is organized as follows. Section \ref{sec: background on CNN and fitler pruning methods} presents a background on CNN and filter pruning methods used to compress CNNs.
Section \ref{sec: proposed pruning framework} describes the proposed filter pruning method. Experimental setup including dataset and unpruned CNNs used for experiments are described in Section \ref{sec: experimental setup}. Section \ref{sec: results} presents results and analysis. Finally, discussion and conclusion is presented in Section \ref{sec: discussion} and \ref{sec: conclusion} respectively.

\begin{figure*}[t]
	\centering
	\includegraphics[scale=0.5]{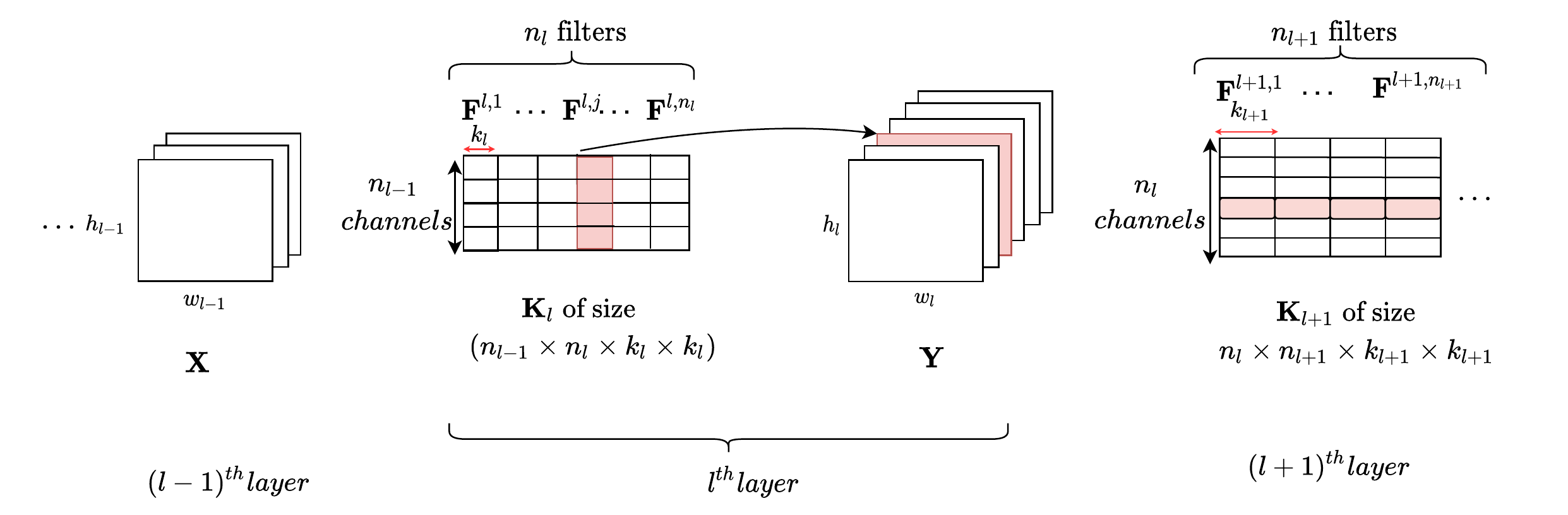}
      \vspace{-0.5cm}
	\caption{An illustration of the intermediate layer structure and filter pruning in  CNN.   In the $l^{\textrm{th}}$ layer, there are $n_l$ filters each having ($k_l \times k_l$) size and $n_{l-1}$ channels. The $n_l$ filters produce $n_l$ feature maps. Pruning the $j^{\text{th}}$ filter, $\mathbf{F}^{l,j}$, in the $l^{\text{th}}$ convolution layer results in elimination of the feature map produced by the pruned filter and corresponding channel in the $(l+1)^{\text{th}}$ layer.}
	\label{fig:proposedpruningillustration2-page-1}
\end{figure*}

\section{A background on CNN and filter pruning methods}
\label{sec: background on CNN and fitler pruning methods}
\subsection{CNN background}

 CNNs consist of multiple layers of filters that are applied to an input image/data to extract different features at different levels of abstraction. Let a CNN has $L$ convolutional layers with indexes $\in$ $\{1,2,\dotsi,l,\dotsi,L\}$. Each convolutional layer consists of a set of 3D convolutional filters having number of channels, length and width as parameters. The convolutional filters produce feature maps after performing convolutional operation on input.  A pictorial illustration of two intermediate layers in the CNN  is described in Figure \ref{fig:proposedpruningillustration2-page-1}.

Let  $n_{l-1}$ denotes the number of  input channels  and ($h_{l-1} \times w_{l-1}$) denotes the size of input feature maps for the $l^{th}$ convolutional layer. Let  $\mathbf{X}$ of size  $(n_{l-1} \times h_{l-1} \times w_{l-1})$ and  $\mathbf{Y}$ of size   $(n_l \times h_{l} \times w_{l})$, denote the  feature map matrices, produced by stacking all the respective feature maps  in the $(l-1)^{\textrm{th}}$ and the $l^{\textrm{th}}$ layer respectively. An element of  $j^{\text{th}}$ feature map in the $l^{\text{th}}$ layer is produced by a convolution operation as given in Equation \ref{Equ: convo operation}, using a $j^{\text{th}}$ filter $\mathbf{F}^{l,j}$ of size ($n_{l-1} \times k_l \times k_l$) having $n_{l-1}$ 2D kernels each of ($k_l \times k_l$) and a sliding window $x$ of size ($n_{l-1} \times k_l \times k_l$) taken from $\mathbf{X}$.  Subsequently a bias and a non-linear activation function is applied on feature maps.

All filters in the $l^{\text{th}}$ convolution layer constitute a kernel tensor $\mathbf{K}_l = [ \mathbf{F}^{l,1}~  \mathbf{F}^{l,2}~ \dots ~\mathbf{F}^{l,n_l}]$ of size $ (n_{l-1} \times n_{l}  \times k_l \times k_l)$.

\begin{equation}
    y = \sum_{c=1} ^{n_{l-1}}\sum_{k_1=1}^{k_l} \sum_{k_2=1}^{k_l}  (x_{c,k_1,k_2} \times \mathbf{F}_{c, k_1, k_2}^{l,j} ) .
    \label{Equ: convo operation}
\end{equation}



\subsection{Filter pruning methods}
\label{sec: filter pruning methods}
Given a trained CNN, pruning filters comprise of two steps.
First, the importance of the convolutional filters is computed and a pruned CNN is obtained after eliminating less important convolutional filters from the original CNN. Second, the pruned network is fine-tuned to regain the most of the performance lost due to filter pruning.  A good survey on such pruning techniques can be found in \cite{liang2021pruning}. 
To compute the importance of the filters, active or passive filter methods have been employed. 
Below, we provide a relevant literature on active and passive filter pruning methods.

\textbf{Active filter pruning:} Active filter pruning methods involve a dataset to compute importance of the filters. For example,  Luo et al. \cite{luo2017entropy}, Lin et al. \cite{lin2020hrank}, and Yeom et al. \cite{yeom2021toward} proposed feature-map based pruning methods, where a dataset is used to produce feature maps in CNNs, and then metrics such as entropy, variance, average rank of feature maps and the average  percentage of zeros are applied on the feature maps to quantify the importance of the filters.

Other methods \cite{liu2017learning,lin2019towards} compute the importance of filters during training process by jointly learning the pruning configuration by associating extra parameters such as a soft mask to each feature map or parameters of the batch normalization layer, and the corresponding feature maps having soft mask value close to zero  are eliminated at the end of the training. 
However, these methods \cite{liu2017learning,lin2019towards}  have significant computational overhead adding upto 10 times more training time \cite{lagunas2021block,xia2022structured}, use extra memory resources to obtain feature maps and involve complex training procedure to optimize  extra parameters such as a soft mask. Also, the pruning method proposed by \cite{liu2017learning} depends whether there exists batch normalization layer in CNN or not. 

\textbf{Passive filter pruning:} On the other hand, passive filter pruning methods \cite{li2016pruning,he2019filter} only use the parameters of the filters without involving any dataset to compute the importance of the filters. Therefore, the passive filter pruning methods are typically less time-consuming and require less memory resources to compute the important filters. In particular, when a pre-trained  network exists, identifying important set of filters using optimization process \cite{liu2017learning,lin2019towards} would be heavily computationally expensive compared to that of the passive filter pruning methods. A typical passive filter pruning method uses an entry-wise norm of the filter ($\mathbf{F}$) to measure the filter importance. For example, an $l_1$-norm, $||\mathbf{F}||_1 =  \sum_{i=1}^{\text{length}(\mathbf{F})}{|\mathbf{F}_i|}$ or an $l_2$-norm, $||\mathbf{F}||_2 =  (\sum_{i=1}^{\text{length}(\mathbf{F})}{\mathbf{F}_i^2})^{\frac{1}{2}}$.
Li et al. \cite{li2016pruning} eliminates filters having smaller entry-wise $l_1$-norm or $l_2$-norm, and finds that eliminating filters based on the entry-wise  $l_1$-norm or $l_2$-norm of the filters gives similar performance. He et al. \cite{he2019filter} eliminates the filters with smaller $l_2$-norm as measured from the  geometric median of all filters.

\subsection{Problems with passive filter pruning methods}
\label{sec: problems with passive filter pruning}

\begin{figure}[t]
    \centering
    \includegraphics[scale=0.47]{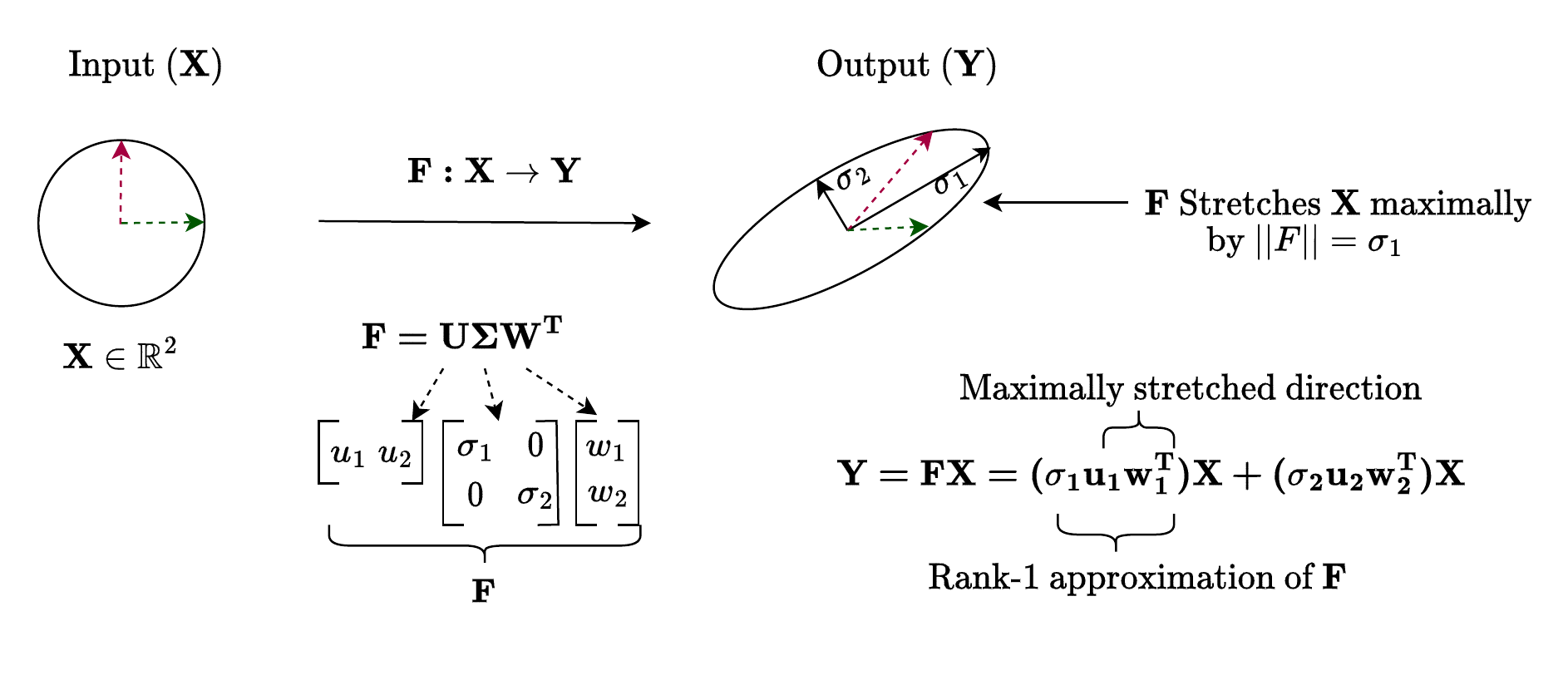}
    \caption{A geometrical view of output produced by a convolution operation, where input feature maps $\mathbf{X}$ in $\mathbb{R}^2$ are transformed to output feature maps $\mathbf{Y}$ in $\mathbb{R}^2$ using a transformation matrix $\mathbf{F}$.  $\mathbf{F}$ is decomposed to a left singular matrix ($\mathbf{U}$), a right singular matrix ($\mathbf{W}$) and a diagonal matrix  ($\Sigma$) using a singular value decomposition method.  $\mathbf{U}$ and $\mathbf{W}$ are orthogonal matrices that cause rotation in the input, and  $\sigma_1$ and $\sigma_2$ are singular values that scale the input. $\mathbf{F}$ stretches $\mathbf{X}$ maximally by $||\mathbf{F}|| = \sigma_1$ which is an operator norm of $\mathbf{F}$.}
    \label{fig: input output illustration}
\end{figure}

However, the passive filter pruning methods assume that a filter with smaller entry-wise norm is less informative or less important, without considering how significantly a filter contribute in producing output. For example, an illustration of the contribution by the filters in producing output is shown in Figure \ref{fig: input output illustration}, where filters $\mathbf{F}$ produces an output $\mathbf{Y}$ by maximally stretching the input $\mathbf{X}$ by a largest singular value $\sigma_1$ that represents an operator norm of  the $\mathbf{F}$.  However, the entry-wise norm methods do not consider any input-output relationship information and rely on each entry of the filter while computing the filter importance.

\begin{figure}[h]
    \includegraphics[scale=0.24]{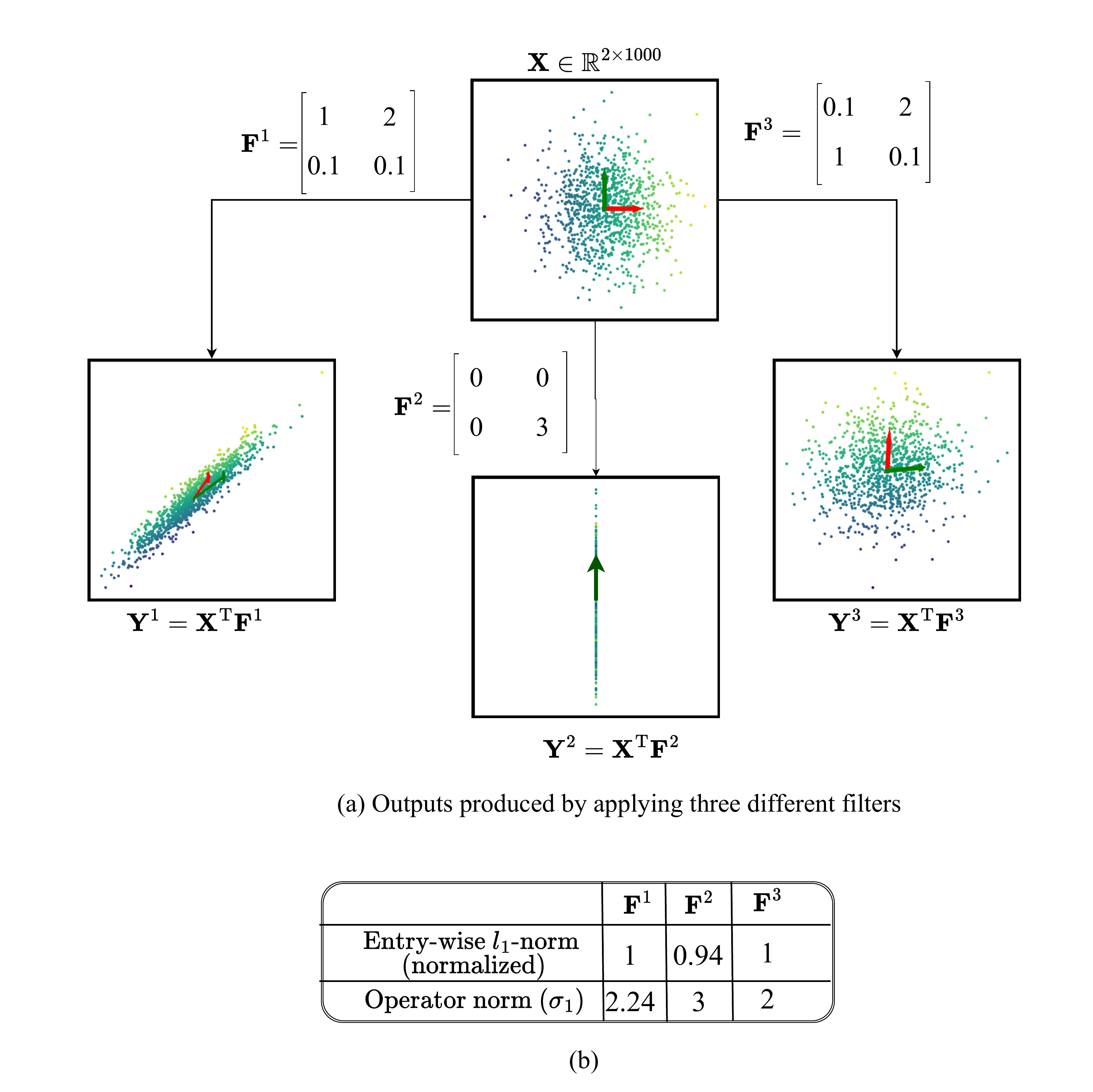}
    \caption{(a) An illustration of output produced in the convolution layer by three CNN filters, $\mathbf{F}^1$, $\mathbf{F}^2$ and $\mathbf{F}^3$ after applying a convolution operation on $\mathbf{X} \in \mathbb{R}^{2 \times 1000}$. (b)  shows the entry-wise $l_1$-norm and the operator norm of the three filters.}
    \label{fig: significance illustration}
\end{figure}

To illustrate the above further, we pictorially show in Figure \ref{fig: significance illustration}(a) that two filters $\mathbf{F}^1$ and $\mathbf{F}^3$ having same entry-wise norm contribute differently and produce different output due to different operator norm of the each filter shown in Figure \ref{fig: significance illustration}(b). Hence such filters should be given different importance. Moreover, when a small number of filters have to be retained in the CNN to yield a very small pruned CNN,  selecting filters with only high entry-wise norm may ignore the smaller norm filters that may also contribute significantly in producing output \cite{ye2018rethinking}. This may degrade the accuracy of the pruned network  significantly due to relying on the filters having  high entry-wise norm, however low significance in producing output. 
For example as shown in Figure \ref{fig: significance illustration}(b), $\mathbf{F}^2$ has $\sigma_1$ = 3, and it stretches $\mathbf{X}$ relatively larger than that of the $\mathbf{F}^1$ and  the $\mathbf{F}^3$. However, the entry-wise norm of the   $\mathbf{F}^2$ is  the smallest among all three filters. Therefore, $\mathbf{F}^2$ shall be given least importance  by the entry-wise norm methods despite a relatively high contribution compared to $\mathbf{F}^1$ and  $\mathbf{F}^3$ filters.

To select the important filters based on their contribution in producing output but without access to data, we propose a novel passive filter pruning framework by considering filters in a convolution layer as an operator that transform input feature maps to output feature maps  rather than relying on  $||\mathbf{F}||_1$ or  $||\mathbf{F}||_2$, the entry-wise norm of the filters.
Utilising all filters in a convolutional layer, we use a Rank-1 approximation of the filters to obtain the maximally stretched direction associated with the operator norm  ($\sigma_1$)  in which the input gets stretched maximally by all filters.
A filter in the convolutional layer is deemed important based on how well it is aligned along the maximally stretched direction represented by all filters in that convolutional layer.


\section{Proposed Operator Norm based Filter Pruning Framework}
\label{sec: proposed pruning framework}


\subsection{Computing filter importance in the $l^{\textrm{th}}$ convolution layer}

Given  a kernel tensor $\mathbf{K}_l$ for the $l^{\textrm{th}}$ convolution layer, our aim is to compute the importance of each of the $n_l$ filters. We transform the kernel tensor  $\mathbf{K}_l$ to  $\mathbf{K^v}_l$ of size $ (n_{l-1} \times n_{l}  \times {k_l}^v)$ by vectorization the 2D kernels each of ($k_l \times k_l$) to ${k_l}^v$.

\begin{figure}[t]
    \centering
	\includegraphics[scale=0.3]{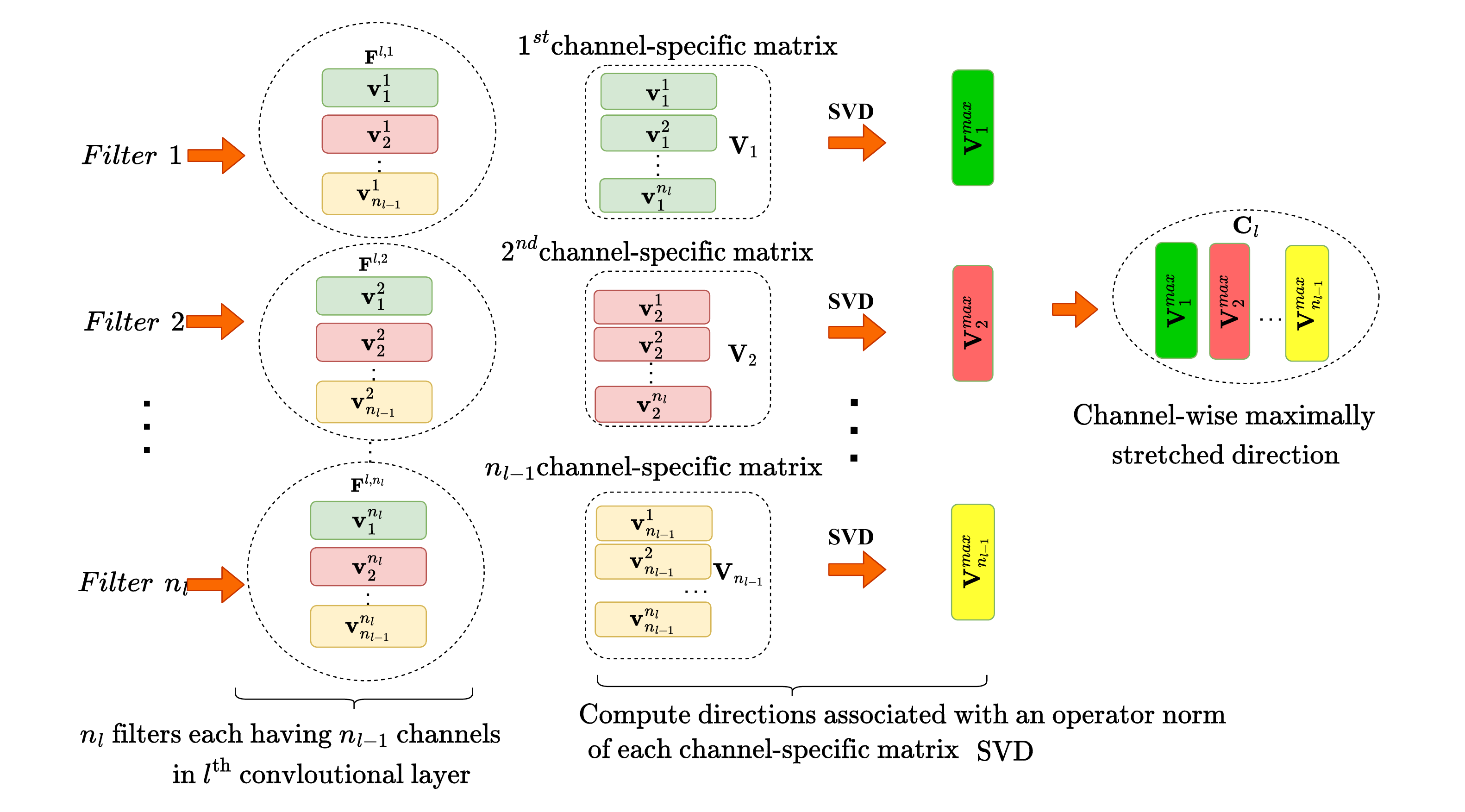}
    \vspace{-0.8cm}
	\caption{An illustration to obtain maximally stretched direction for each $n_{l-1}$ channels across all filters in the $l^{th}$ convolutional layer.}
	\label{fig:csproposednewall}
\end{figure}

As the output produced by the $c^{\textrm{th}}$ channel of a filter is associated with only the $c^{\textrm{th}}$ channel of the input in CNNs, we obtain maximally stretched direction of each channel across filters independently. An illustration of the procedure to obtain maximally stretched directions for each channels is shown in Figure \ref{fig:csproposednewall}. 

A channel-specific matrix $\mathbf{V}_c$ of size (${n_l \times {k_l}^v}$) is constructed by taking the $c^{\textrm{th}}$ channel of all filters.
$\mathbf{V}_c$ denotes the learned weights of the $c^{\textrm{th}}$ channel across the $n_l$ filters.  Next, singular value decomposition (SVD) is performed on $\mathbf{V}_c$ to compute a rank-1 approximation of the $c^{\textrm{th}}$ channel as $\mathbf{V}_c \approx \sigma^c_1 \mathbf{u}^c_1 {\mathbf{w}^c_{1}}^\text{T}$. Here,  $\sigma^c_1$ denotes the maximum singular value that  affects the corresponding channel of the input maximally and is equivalent to the operator norm of the $\mathbf{V}_c$, $\mathbf{u}^c_1$ is the first left singular vector, and $\mathbf{w}^c_1$ is the first right singular vector.

A row  of $\mathbf{V}_c$ normalized to unit norm is denoted as $\mathbf{v}_c^{\text{max}}$, and is considered as a  maximally stretched direction for the $c^{\textrm{th}}$ channel.
The $\mathbf{v}_c^{\text{max}}$ provides a reference for measuring the significance the $c^{\textrm{th}}$ channel of the filter.

After obtaining $\mathbf{v}_c^{\text{max}}$ corresponding to the  $c^{\textrm{th}}$ channel, we obtain maximally stretched directions  for other channels. Finally, the $\mathbf{v}_c^{\text{max}}$ of all channels are  stacked together to yield $\mathbf{C}_l$ of size (${{k_l}^v \times n_{l-1}}$) for a given convolutional layer.


Given $\mathbf{C}_l$, the importance for the $j^{\textrm{th}}$ filter  is computed as, $\text{Trace}[(\mathbf{F}^{l,j})\mathbf{C}_l)]$.  The filters are ranked as per their importance with a relatively high importance score indicates a high significance of the filter in producing output. Algorithm \ref{alg: PCS algo} summarizes the process to compute the importance  of various filters in a given convolutional layer.

After ranking the filters based on their importance for various convolutional layers, few unimportant filters are eliminated based on a user-defined pruning ratio for various convolutional layers of CNN, and  a pruned network is obtained. Pruning a filter in the $l^{\text{th}}$ convolution layer also eliminates the feature map produced by the pruned filter  and the related kernel or channel in the next layer as shown in Figure  \ref{fig:proposedpruningillustration2-page-1}, hence the computations is reduced in both the $l^{\text{th}}$ layer and the $(l+1)^{\text{th}}$ layer. In the end, fine-tuning of the pruned network is performed to regain some of the performance loss due to  elimination of filters from the original network.

\begin{algorithm}[t]
	\SetAlgoLined
	\caption{Filter Importance Calculation}\label{alg: PCS algo}
	\SetKwData{Left}{left}
	\SetKwData{This}{this}
	\SetKwData{Up}{up}
	\SetKwFunction{Union}{Union}
	\SetKwFunction{FindCompress}{FindCompress}
	\SetKwInOut{Input}{input}
	\SetKwInOut{Output}{output}
	\Input{$\mathbf{K^v}_l$ of size $(n_{l-1} \times n_l \times {k_l}^v)$, \textrm{kernel matrix in  $l^{\textrm{th}}$ layer }.
	}
	\Output{Normalized importance scores}   
	Initialization: $\mathbf{C}_l$ = [ ],
	Score = [ ],  \\
	...
	\textcolor{blue}{\enquote{Obtaining channel-wise maximally stretched direction associated with operator norm}}...\\
	\For{$c\gets1$ \KwTo $n_{l-1}$ }{
		$\mathbf{V}_c$ = $\mathbf{K^v}_l$[c , : , :] \Comment{\textcolor{blue}{Take $c^{\textrm{th}}$ channel of all the \hspace*{3cm} filters}} \\
		\hspace*{-0.2cm}$\mathbf{u}^{c}$, $\Sigma^{c}$,$\mathbf{w}^{c}$ = SVD( $\mathbf{V}_c$ ) \Comment{\textcolor{blue}{Perform SVD on  $\mathbf{V}_c$}}\\
		\hspace*{-0.2cm}$\mathbf{C}_l$.append(($\mathbf{u}_1^{c}$ $\mathbf{w}_1^{c^{\text{T}}}$)[1,:]) \Comment{\textcolor{blue}{$c^{th}$ channel maximally \hspace*{2.2cm}  stretched directions related to $\sigma_1$.}}
  
	 }

....................\textcolor{blue}{ \enquote{Importance score calculation}}.................... \\
	\For{$j\gets1$ \KwTo $n_l$}{
		$\mathbf{F}^{l,j}$ = $\mathbf{K^v}_l$[ : , j , :] \Comment{\textcolor{blue}{Take $j^{\textrm{th}}$ filter.}} \\
		\hspace*{-0.2cm}$\mathbf{\bar{F}}^{l,j}$= \textrm{Score.append}$($[trace($(\mathbf{F}^{l,j})\mathbf{C}_l$)$])$ \Comment{\textcolor{blue}{Compute  \hspace*{5cm} importance.}} 
	}
	$\alpha$ =$[\textrm{Score}]$ 
     \\
	\hspace*{-0.1cm}Score\_normalized        =   $\frac{\alpha^2}{\textrm{max}(\alpha^2)}$ \Comment{\textcolor{blue}{Normalized importance \hspace*{4.2cm} scores.}}
	
%
 \noindent  return Score\_normalized \label{Algo: proposed filter pruning}
	
\end{algorithm}

\section{Experimental Setup}
\label{sec: experimental setup}

We evaluate the proposed pruning method on CNNs designed for audio scene classification (ASC) and image classification. We use etiher publicly available pre-trained network or we train the networks from scratch to obtain an unpruned CNN. A brief summary of the unpruned CNNs used for experimentation is shown in Figure \ref{fig: unpruned CNNs for experiments} and is described below

\textbf{Unpruned CNNs:}  We use two different unpruned networks for ASC, (a)  VGGish\_Net and (b) DCASE21\_Net. We also use (c) VGG-16 network and (d) ResNet-50 for image classification.

\begin{enumerate}[label=(\alph*)]
    \item \textbf{VGGish\_Net:} The VGGish\_Net is built using a publicly available pre-trained VGGish network \cite{VGGish} followed by a dense and a classification layer. The VGGish\_Net has six convolution layers (termed as C1 to C6). We train the VGGish\_Net on TUT Urban Acoustic Scenes 2018 development (we denote the dataset  as DCASE-18) training dataset \cite{Mesaros2018_DCASE} to classify 10 different audio scenes using an Adam optimizer with cross-entropy loss function for 200 epochs. Each audio recording in DCASE-18 dataset has 10s length.
    
    The input to the VGGish\_Net is a log-mel spectrogram of size 96 $\times$ 64 computed corresponding to 960ms audio segment. The VGGish\_Net has 55,361,162 (approximately 55M) parameters, requiring 903M multiply-accumulate operations (MACs)\footnote{\textcolor{black}{To compute MACs, we use publicly available \enquote{nessi.py} script available at \url{https://github.com/AlbertoAncilotto/NeSsi}}} during inference corresponding to an audio clip of 960ms and the trained network gives  64.69\% accuracy on 10s audio scene for DCASE-18 development validation dataset. 

    \item \textbf{DCASE21\_Net:} DCASE21\_Net is a publicly available pre-trained network designed for DCASE 2021 Task 1A that is trained using TAU Urban Acoustic Scenes 2020 Mobile development dataset (we denote \enquote{DCASE-20}) to classify 10 different audio scenes \cite{martin2021low}. DCASE21\_Net consists of three convolutional layers (termed as C1 to C3) and one fully connected layer. The input to the network is a log-melspectrogram of size (40 × 500) corresponding to a 10s audio clip. The trained network  has 46246 parameters, requiring approximately 287M MACs during inference corresponding to  10-second-length audio clip and gives 48.58\% accuracy on the DCASE-20 development validation dataset.

\item \textbf{VGG-16:} We use a VGG-16 network \cite{liu2015very} for handwritten digit classification and CIFAR-10 classification \cite{krizhevsky2009learning}. The VGG-16  consists of 13 convolutional layer (termed as C1 to C13) and 2 dense layers. The VGG-16 has 15M parameters and requires 329M MACs per inference corresponding to an input of size (32 $\times$ 32 $\times$ 3).

\begin{figure}[t]
    \centering
    \includegraphics[scale=0.5]{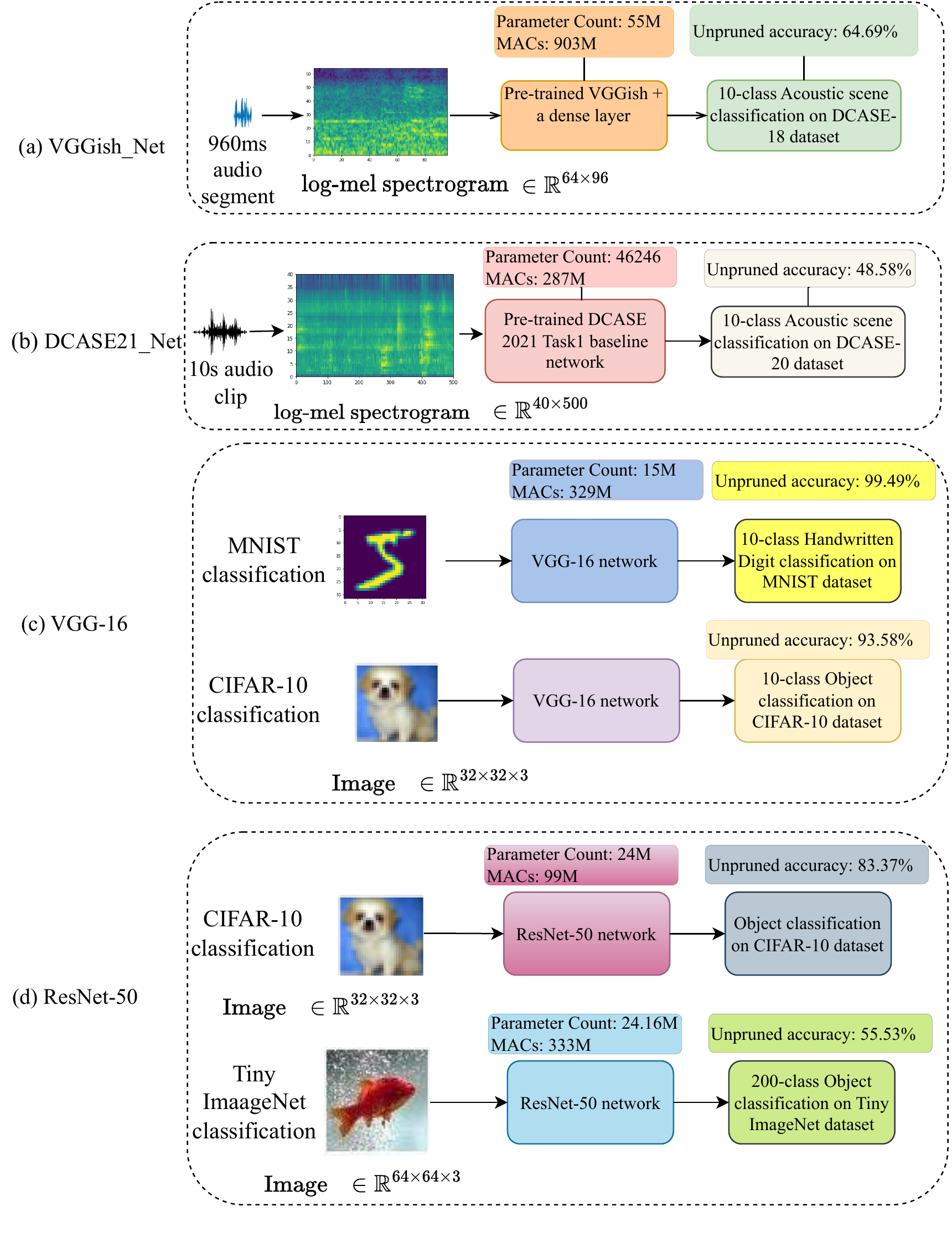}
    \caption{Unpruned CNNs used for experimentation; (a) VGGish\_Net (b) DCASE21\_Net, (c) VGG-16 and (d) ResNet-50.}
    \label{fig: unpruned CNNs for experiments}
\end{figure}

\textbf{(i) VGG-16 on MNIST:} For handwritten digit classification, we use MNIST dataset\footnote{The MNIST dataset is downloaded using Keras API which has a pre-defined training and testing set of grayscale images. Each grayscale image of size (28 $\times$ 28)  is converted into three channels by depthwise stacking the grayscale image and then reshaping to  (32 $\times$ 32 $\times$ 3) that is used as an input to VGG-16.}\cite{lecun2010mnist}. We train the VGG-16 from scratch for 200 epochs on the MNIST training dataset and the trained VGG-16 gives 99.49\% accuracy for the MNIST testing dataset.

 \textbf{(ii) VGG-16 on CIFAR-10:}  We use a publicly available pre-trained VGG-16 network \cite{cifar_vgg,liu2015very} trained on CIFAR-10 dataset as an unpruned network.  The  pre-trained VGG-16 gives 93.58\% accuracy for the CIFAR-10 testing dataset. 

\item \textbf{ResNet50\_Net:}  We use a ResNet-50 architecture pre-trained with ImageNet \cite{he2016deep} weights followed by a global average pooling, a fully connected layer and a classification layer to train an unpruned network for (i) CIFAR-10 and (ii) Tiny ImageNet \cite{wu2017tiny} dataset classification.  The Tiny ImageNet dataset \cite{wu2017tiny} is a subset of the ImageNet Large Scale Visual Recognition Challenge (ILSVRC) dataset \cite{russakovsky2015imagenet}. The Tiny ImageNet consists of  200 different image classes with each image is re-sized to (64x64) pixels from (256x256) pixels in standard ImageNet. There are 500 images per class for training and 50 images per class for validation.  We download the Tiny ImageNet dataset from \cite{Tiny_imagenet_data}. 

To obtain the unpruned network, we train ResNet50\_Net for 300 epochs with stochastic gradient descent (SGD) optimizer and cross entropy loss function.  ResNet50\_Net has five stages and each stage comprises of convolutional layers across the main branch and  the residual branch. An overall architecture of ResNet50\_Net is shown in Figure \ref{fig: resnet architecture for imagenet classifcaiton} in the supplementary material.

\textbf{(i) ResNet50\_Net on CIFAR-10:} For CIFAR-10, the unpruned ResNet50\_Net has 24M parameters, 99M MACs and gives 83.37\% accuracy on CIFAR-10 test dataset with (32 x 32 x 3) shaped input.
 
\textbf{(ii) ResNet50\_Net on Tiny ImageNet:} For Tiny ImageNet classification, the unpruned ResNet50\_Net has 24.16M parameters, 333M MACs per inference with (64 x 64 x 3) shaped input, and achieves 55.53\% accuracy similar to that obtained by Sun \cite{sun2016resnet} on validation dataset. 

\end{enumerate}

\textbf{Obtaining pruned network and performing fine-tuning:} After computing importance of the filters using Algorithm \ref{Algo: proposed filter pruning} across various convolutional layers, we eliminate $p$ percentage  of unimportant filters from various convolutional layer, where $p$ $\in$ \{25, 50, 75, 90\} denotes a pruning ratio. For plain CNNs, (a), (b) and (c) as shown in Figure \ref{fig: unpruned CNNs for experiments}, we prune convolutional filters from subset of convolutional layers and obtain subset of pruned networks. For residual CNN (d), as shown in Figure \ref{fig: unpruned CNNs for experiments}, we prune convolutional filters across various stages of the network to obtain various pruned networks.

Once the pruned network is obtained, we re-train the network with similar conditions such as same optimizer as used in the training of the unpruned networks (a-d) for 50\% fewer epochs compared to the epochs used in training the unpruned network.

\textbf{Other methods used for comparison:} We compare the proposed operator norm based pruning method with that of the entry-wise norm based methods, (a) $l_1$-norm method that eliminates filters with smaller entry-wise $l_1$-norm \cite{li2016pruning} and (b) geometric median (GM) method that eliminates filters with smaller $l_2$-norm as measured from the geometric median of all filters \cite{he2019filter}. We also compare the proposed pruning method with the existing active filter pruning methods including HRank  \cite{lin2020hrank} and Energy-aware pruning \cite{yeom2021toward}.  The HRank method opts three steps to obtain a pruned network. 1) A set of feature maps are generated for a given filter using a set of examples. 2) Rank of the feature maps is computed and an average rank computed for various examples is used as a criterion to quantify filter importance. 3) The filters with low average rank are eliminated and a fine-tuning procedure is opted to compensate the performance loss. Similar to the HRank method, the Energy-aware method uses a set of input data to generate feature maps for a given convolutional layer and then compute energy of each feature map by computing nuclear norm (sum of all singular values) of each feature map. A feature map with low energy is pruned and then fine-tuning of the pruned network is performed. To perform HRank and Energy-aware pruning, we randomly select 500 examples from the training dataset to generate feature maps.


\textbf{Performance metrics:} We analyse accuracy, the number of MACs and the number of parameters in the pruned networks obtained after pruning various subsets of convolution layers at different pruning ratios from the unpruned network. We also analyse the computational time required to compute the importance of filters to compare with the active filter pruning methods.

\section{Results and Analysis}
\label{sec: results}
\subsection{Analysing MACs, parameters and accuracy after pruning various subset of convolutional layers/stages}

\begin{figure}[t]
    \centering
    \includegraphics[scale=0.56]{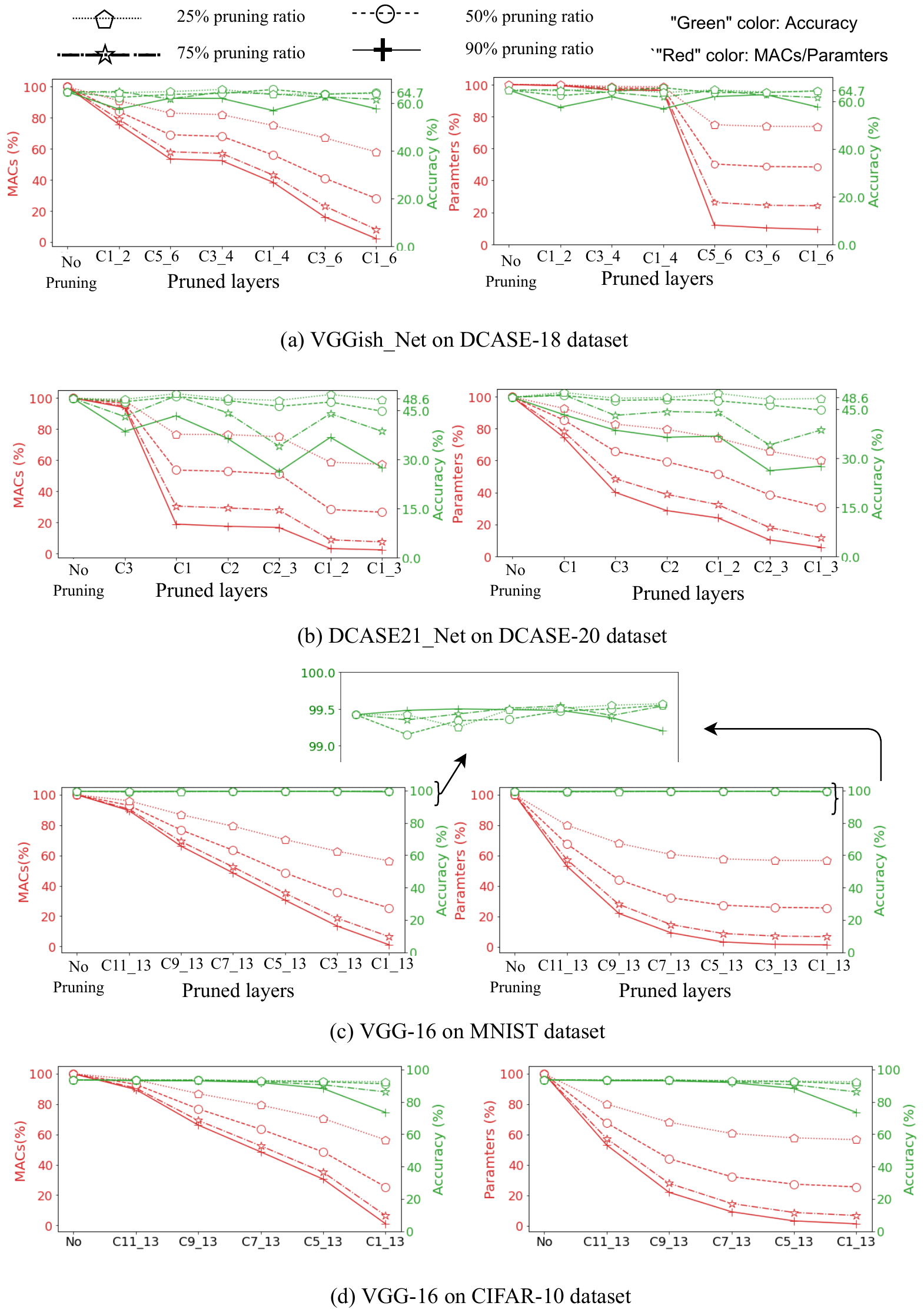}
    \vspace{-0.5cm}
    \caption{Accuracy, MACs and parameters across different pruned networks for (a) VGGish\_Net, (b) DCASE21\_Net and (c) VGG-16 network on MNIST and (d) VGG-16 on CIFAR-10, when different subsets of convolutional layers are pruned at different pruning ratios. Here, \enquote{CA\_B} as \enquote{pruned layers} means that convolution layers from A to B are pruned.}
    \label{fig: ASC layer wise analysis}
\end{figure}

Figure \ref{fig: ASC layer wise analysis} shows accuracy, the number of parameters and the number of MACs obtained after pruning various subset of convolutional layers from the unpruned networks: VGGish\_Net, DCASE21\_Net and VGG-16.

\noindent \textbf{VGGish\_Net and DCASE21\_Net on ASC:} As shown in Figure \ref{fig: ASC layer wise analysis} (a) and \ref{fig: ASC layer wise analysis} (b), 
the accuracy obtained using the various pruned networks is similar to that of the unpruned networks at $p =$ 25\%. For VGGish\_Net,  the number of MACs are reduced by 40 percentage points and the parameters are reduced by 25 percentage points, when all convolutional layers (C1\_6: C1 to C6) are pruned at $p =$ 25\% shown in Figure \ref{fig: ASC layer wise analysis}(a). For DCASE21\_Net, both the MACs and the parameters are reduced by 40 percentage points when all convolutional layers (C1\_3: C1 to C3) are pruned at $p =$ 25\% shown in Figure \ref{fig: ASC layer wise analysis}(b).

At $p =$ 50\%, the accuracy drop across various pruned networks is less than 4 percentage points compared to that of the unpruned network for both VGGish\_Net and DCASE21\_Net as shown in Figure \ref{fig: ASC layer wise analysis}(a), (b). For VGGish\_Net,  the number of MACs are reduced by 75 percentage points and the parameters are reduced by 55 percentage points, when all convolutional layers (C1\_6: C1 to C6) are pruned at $p =$ 50\% shown in Figure \ref{fig: ASC layer wise analysis}(a). For DCASE21\_Net, both the MACs and the parameters are reduced by 75 percentage points when all convolutional layers (C1\_3: C1 to C3) are pruned at $p =$ 50\% shown in Figure \ref{fig: ASC layer wise analysis}(b). 

At $p =$ 75\%, the accuracy drop across various pruned networks is less than 5 percentage points and 10 percentage points compared to that of the unpruned network for both VGGish\_Net and DCASE21\_Net respectively as shown in Figure \ref{fig: ASC layer wise analysis}(a), (b). The accuracy of the pruned networks degrades further at $p =$ 90\% and the drop in accuracy for VGGish\_Net and DCASE21\_Net is  10 percentage points and 20 percentage points  respectively, when all layer are pruned at $p$ = 90\%. On the other hand, both the MACs and the parameters are reduced significantly by more than 75 percentage points when all convolutional layers of VGGish\_Net and DCASE21\_Net are pruned at $p = \{75\%, 90\%\}$  as shown in Figure \ref{fig: ASC layer wise analysis}(a), (b).


In general, the MACs, the parameters and the accuracy decrease when various convolutional layers are pruned from 25\% to 90\% pruning ratio. The accuracy  of the pruned DCASE21\_Net  reduces significantly
from 0.2 percentage points to 20 percentage points compared to that of the unpruned network, when all layers are pruned with 25\% to 90\% pruning ratio. This might be due to the smaller network  size of the DCASE21\_Net, where eliminating large number of parameters at high pruning ratio results in under-fitting problem due to insufficient parameters.

\noindent \textbf{VGG-16 for image classification :} For VGG-16 on MNIST as shown in Figure \ref{fig: ASC layer wise analysis}(c), the accuracy of the various pruned networks is reduced by less than 0.5\% percentage points compared to that of the unpruned VGG-16 . The number of MACs and the parameters are reduced from approximately 40 percentage points to 99 percentage points when pruning ratio across all convolutional layers (C1\_13: C1 to C13)  varies from 25\% to 90\% respectively. Therefore, we reduce the computational complexity of the unpruned VGG-16 for MNIST classification at approximately no loss of accuracy.

For VGG-16 on CIFAR-10 as shown in Figure \ref{fig: ASC layer wise analysis}(d), varying $p$  from 25\% to 70\% across convolution layers from C5 to C13, the number of parameters are reduced from 40\% to 90\%, and the  MACs are reduced from 25\% to 70\% with an accuracy drop from 0.25 to 3 percentage points respectively. Pruning all convolutional layers at $p = 25\%$ results in approximately 45\% reduction in both parameters and MACs at marginal drop in accuracy compared to that of the unpruned network. On the other hand, pruning all convolutional layers at $p = 90\%$ reduces parameters and MACs significantly, however accuracy drop is approximately 20 percentage points compared to the unpruned network.

Therefore, the pruning ratio can be chosen according to the requirement whether the underlying resources (computation and parameters) are primarily important or the accuracy.

\begin{figure}[t]
    \centering
    \includegraphics[scale=0.43]{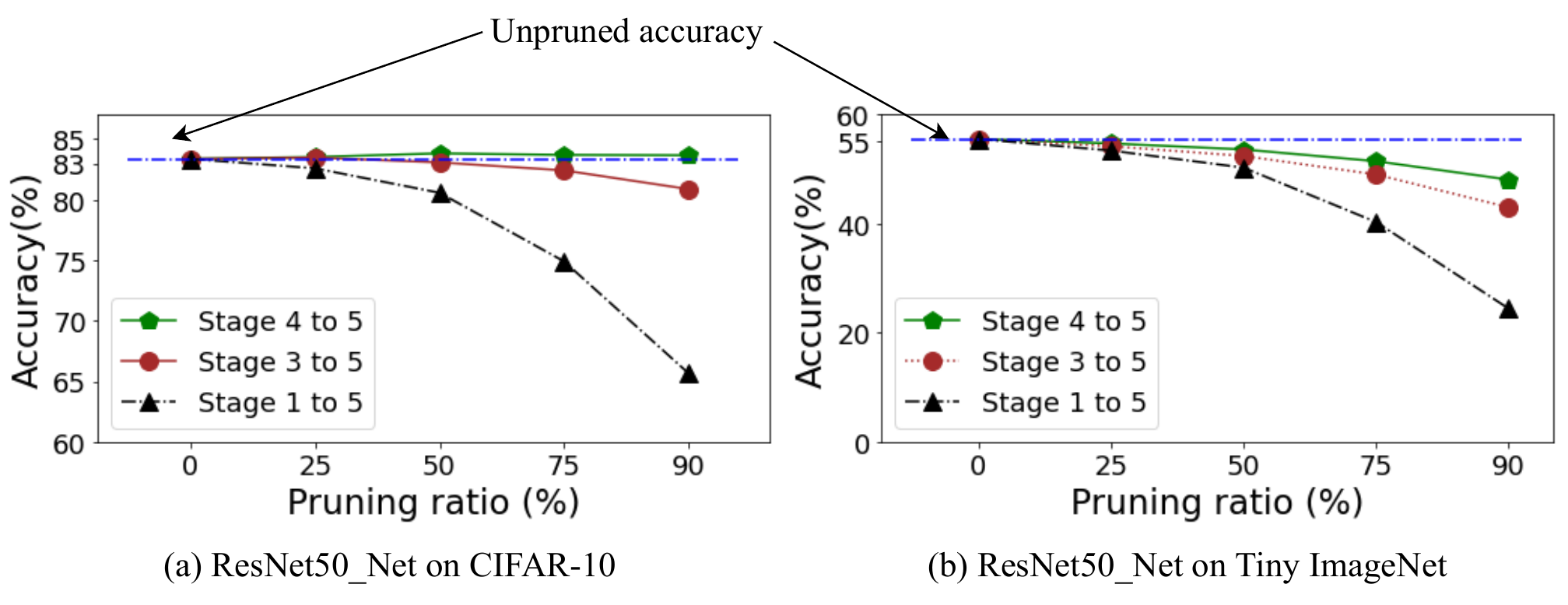}
    \caption{Accuracy obtained at different pruning ratio after pruning convolutional filters from various stages in  ResNet50\_Net for CIFAR10 and Tiny ImageNet datasets.}
    \label{fig: resnet stage wise analysis}
\end{figure}

\begin{figure}[t]
    \centering
    \includegraphics[scale=0.38]{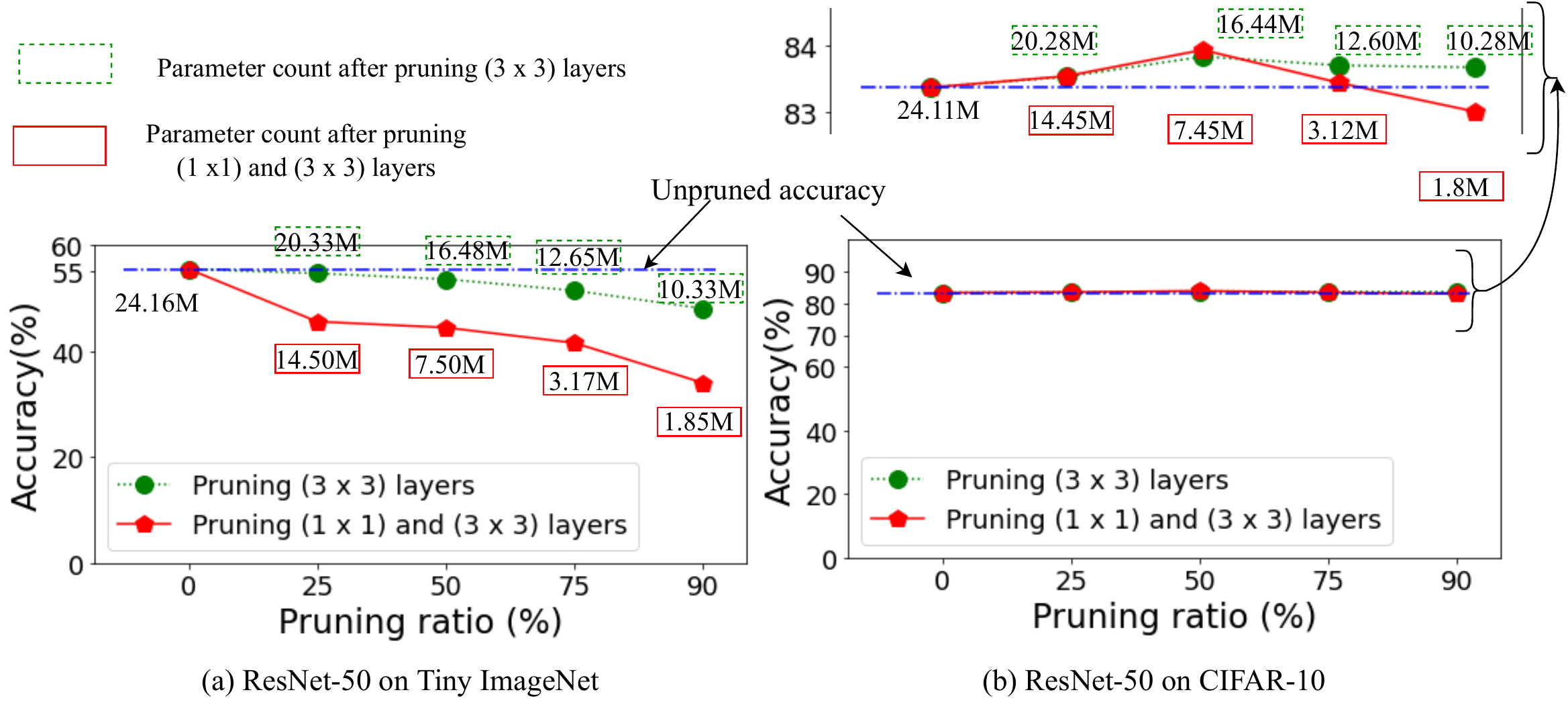}
    \caption{Accuracy obtained after pruning convolutional layers from main branch and residual branch in stage 4 and stage 5 of ResNet50\_Net for Tiny ImageNet and CIFAR-10 datasets.}
    \label{fig: resnet_imagenetpruning different layers comparison}
\end{figure}

\noindent \textbf{ResNet50\_Net for image classification:} Figure \ref{fig: resnet stage wise analysis} shows accuracy obtained in the pruned ResNet\_Net for CIFAR-10 and Tiny ImageNet classification, when convolutional layers from the main branch across different stages are pruned at different pruning ratios. The accuracy drop increases as the number of stages to be pruned and the pruning ratio increases.

Figure \ref{fig: resnet_imagenetpruning different layers comparison} compares the accuracy obtained after pruning  convolutional layers  in the main branch alone, and pruning  convolutional layers across both main  and residual branch in stage 4 and stage 5 of ResNet50\_Net for Tiny ImageNet and CIFAR-10 datasets.  For Tiny ImageNet as shown in Figure \ref{fig: resnet_imagenetpruning different layers comparison}(a), the accuracy obtained after pruning convolutional layers in both main and residual branch is less than that of pruning  convolutional layers in the main branch alone. On the other hand, we obtain similar accuracy for CIFAR-10 dataset after pruning convolutional layers in the main branch alone, and both main and residual branch. This suggests that the ResNet50\_Net shows more redundancy for smaller classification problems with 10 classes having sufficient training examples (5000 per class) compared to that of the relatively difficult classification problem with 200 classes having 500 training examples per class. In addition, there is a little scope to prune convolutional layers in the residual branch alongwith the convolutional layers in the main branch compared to that of pruning only convolutional layers in the main branch alone across various stages in ResNet50\_Net  for Tiny ImageNet classification due to significant drop in accuracy compared to that of the unpruned network as shown in Figure \ref{fig: resnet_imagenetpruning different layers comparison}(a).

\begin{figure}[t]
    \centering
    \includegraphics[scale=0.4]{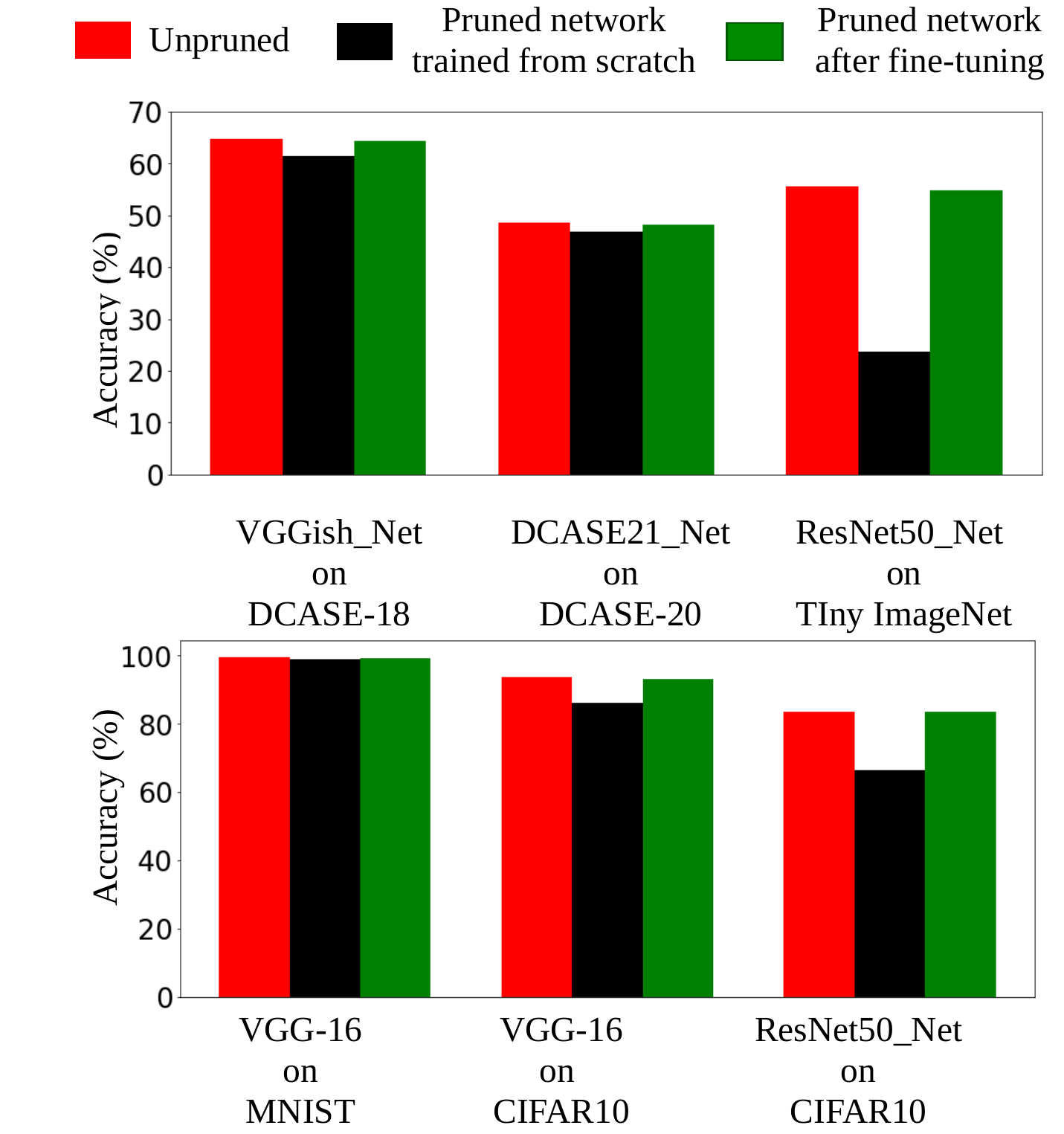}
    \caption{Comparison of accuracy among unpruned network, the pruned network obtained using the proposed pruning method after fine-tuning and the same pruned network obtained using the proposed pruning method, however, trained from scratch.}
    \label{fig: accuracy scrathc versu fien-tuned}
\end{figure}

For various CNNs, Figure \ref{fig: accuracy scrathc versu fien-tuned} compares the accuracy obtained using the pruned network obtained using the proposed pruning method after fine-tuning process with that of the same pruned network obtained using the proposed pruning method, however trained from scratch with random initialization. We find that the accuracy of the pruned network obtained after fine-tuning  is better than  the same pruned network trained from scratch showing the advantage of applying pruning to obtain a smaller size CNN as compared to training smaller size CNN from scratch.

\subsection{Comparison with other methods}

\begin{figure}[t]
    \centering
    \includegraphics[scale=0.32]{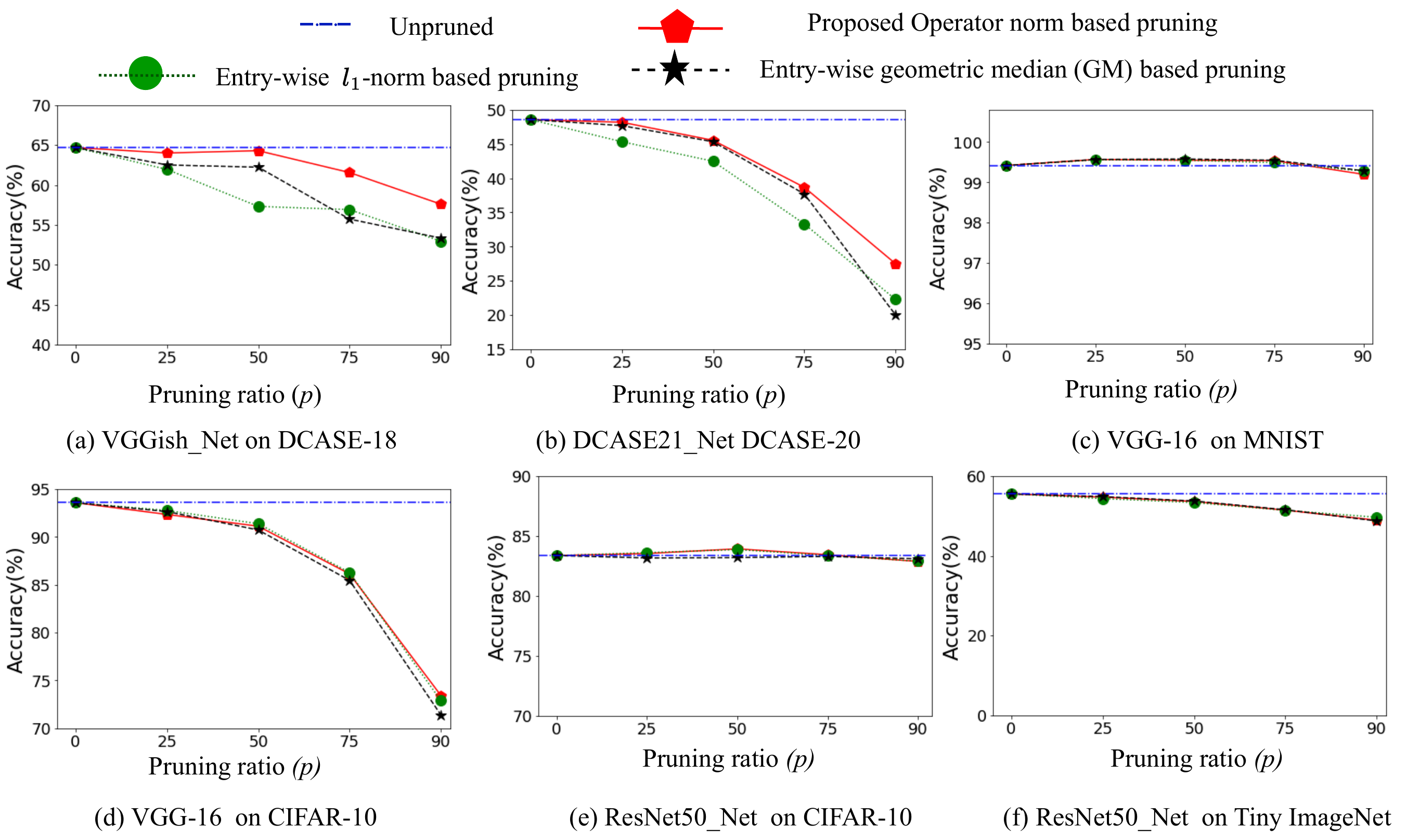}
    \caption{Accuracy obtained at different pruning ratio across various pruned networks obtained using the proposed pruning method, entry-wise $l_1$-norm and geometric median (GM) based pruning methods.}
    \label{fig: comparison with passive}
\end{figure}

Figure \ref{fig: comparison with passive} compares the accuracy of the proposed operator norm pruning method with that of the entry-wise $l_1$-norm and the geometrical median (GM) based methods, when the unpruned network is pruned at different pruning ratios. For audio classification networks including VGGish\_Net and DCASE21\_Net, the accuracy obtained using the proposed operator norm based pruning method is better than that of the entry-wise norm methods particularly when a large number of filters ($p = 90\%$) are pruned from the network as shown in Figure \ref{fig: comparison with passive}(a), (b). On the other hand, for image classification networks such as VGG-16 and ResNet50\_Net, the  accuracy shown in Figure \ref{fig: comparison with passive}(c)-(h) at various pruning ratio is more or less similar. The convergence plots obtained during the fine-tuning of the pruned networks are included in supplementary material Figure  \ref{fig: convergence plots}


\begin{table*}[t]
	\caption{Comparison of accuracy, the MACs, the parameters  and pruning time of the proposed operator norm based pruning method with existing feature map based pruning methods such as HRank and Energy aware for ResNet-50 on Tiny ImageNet and CIFAR-10 dataset. The pruning time is a local time duration between the start and end of the pruning algorithm in computing the filter importance for all layers in the unpruned network.  The pruning time is computed using python time method time.asctime().}
    \vspace{0cm}
	\centering
     \Large  
	\resizebox{\textwidth}{!}{
		\begin{tabular}{c|cccccccc} \toprule
            {Network} &{Pruning Method} & {Data used in Pruning} & {Pruning Time (Seconds)}  &{Accuracy (\%)}  & {Parameters} & {MACs}\\ \midrule
            {VGGish\_Net} & {No pruning (Baseline)}& {-} & {-}  & {64.69} & {55M} & {903M}  \\
   			{on} & {HRank \cite{lin2020hrank}}& {\ding{51}} & {53}   & {64.35} & {26.82M} & {253M}  \\
            {DCASE-18} & {Energy-aware pruning \cite{yeom2021toward}}& {\ding{51}} & {51}   & {64.38} & {26.82M} & {253M} \\
            {(Input: 64 $\times$ 96 $\times$ 1)} & {Ours} & {\ding{53}} & {0.03} & {64.30}  & {26.82M} & {253M}\\ \midrule
            {DCASE21\_Net} & {No pruning (Baseline)}& {-} & {-}  & {48.58} & {46.246K} & {286M}  \\
   			{on} & {HRank \cite{lin2020hrank}}& {\ding{51}} & {23}   & {48.58} & {27.91K} & {164M} \\
            {DCASE-20} & {Energy-aware pruning \cite{yeom2021toward}}& {\ding{51}} & {21} & {48.10} & {27.91K} & {164M}\\
            {(Input: 40 $\times$ 500 $\times$ 1)} & {Ours} & {\ding{53}} & {0.001}  & {48.18}  & {27.91K} & {164M}\\ \midrule         
            {VGG-16} & {No pruning (Baseline)}& {-} & {-} & {99.49} & {15M} & {329M}  \\
   			{on} & {HRank \cite{lin2020hrank}}& {\ding{51}} & {88} &  {99.16} & {0.18M} & {3.33M}  \\
            {MNIST} & {Energy-aware pruning \cite{yeom2021toward}}& {\ding{51}} & {86} & {99.10} & {0.18M} & {3.33M}\\
            {(Input: 32 $\times$ 32 $\times$ 3)} & {Ours} & {\ding{53}} & {0.08} &  {99.10}  & {0.18M} & {3.33M}\\ \midrule
            {VGG-16} & {No pruning (Baseline)}& {-} & {-} &  {93.56} & {15M} & {329M}  \\
   			{on} & {HRank \cite{lin2020hrank}}& {\ding{51}} & {85} &  {93.16}  & {3.29M} & {217M}  \\
            {CIFAR-10} & {Energy-aware pruning \cite{yeom2021toward}}& {\ding{51}} &  {83} & {93.09}  & {3.29M} & {217M} \\
            {(Input: 32 $\times$ 32 $\times$ 3)} & {Ours} & {\ding{53}} & {0.08} &  {93.00}  & {3.29M} & {217M}\\ \midrule
            
            {ResNet50\_Net} & {No pruning (Baseline)}& {-} & {-} & {55.53} & {24.16M} & {333M}  \\
   			{on} & {HRank \cite{lin2020hrank}}& {\ding{51}} & {531} & {54.82} & {20.32M} & {298M}  \\
            {Tiny ImageNet} & {Energy-aware pruning \cite{yeom2021toward}}& {\ding{51}} & {458} &  {54.60} & {20.32M} & {298M} \\
            {(Input: 64 $\times$ 64 $\times$ 3)} & {Ours} & {\ding{53}} & {113} &  {54.70}  & {20.32M} & {298M}\\ \midrule
            {ResNet50\_Net} & {No pruning (Baseline)}& {-} & {-} & {83.37} & {24.11M} & {99M}  \\
   			{on} & {HRank \cite{lin2020hrank}}& {\ding{51}} & {464} &  {83.60} & {3.12M} & {39M}  \\
            {CIFAR-10} & {Energy-aware pruning \cite{yeom2021toward}}& {\ding{51}} & {427} & {83.43} & {3.12M} & {39M} \\
            {(Input: 32 $\times$ 32 $\times$ 3)} & {Ours} & {\ding{53}} & {113} &  {83.44}  & {3.12M} & {39M}\\
            \bottomrule
 	\end{tabular}}
	\label{tab: ResNet-50 on cifar-10 and imagenetwith active methods  COMPARISON L1 GM PCS}
\end{table*}

Next, Table \ref{tab: ResNet-50 on cifar-10 and imagenetwith active methods  COMPARISON L1 GM PCS} gives various performance parameters including computation time in obtaining filter importance (pruning time), accuracy obtained  after fine-tuning the pruned network, the number of parameters and the number of MACs obtained in various pruned networks after applying  the proposed filter pruning method and the feature map based active filter pruning methods such as HRank \cite{lin2020hrank} and Energy-aware \cite{yeom2021toward} on various unpruned networks.  The proposed passive filter pruning method is able to achieve similar performance compared to the feature map based active filter pruning methods without using any dataset in computing filter importance. Moreover, the computational time in obtaining filter importance using the proposed filter pruning method is significantly less than that of the feature map based active filter pruning methods.

\section{Discussion}
\label{sec: discussion}

In this paper, we prune  convolutional layers in various CNNs with plain and residual architecture designed for audio and image classification by considering operator norm of the filters. We find that the proposed filter pruning method reduces the number of parameters and the number of MACs significantly at marginal reduction in accuracy compared to that of the unpruned network. Considering operator norm of the filter in computing filter importance gives similar performance for CNNs designed for image classification. For audio classification CNNs, the proposed pruning method improves the accuracy compared to that of the entry-wise norm methods at various pruning ratios. Hence, the proposed operator norm based filter pruning method shows better generalization compared to entry-wise norm methods.

For ResNet50\_Net, we find that pruning convolutional layers in the residual branches show more drop in accuracy compared to that of pruning convolutional layer in the main branch alone for Tiny ImageNet classification having 200 classes. On the other hand, pruning convolutional layers in the residual branches for CIFAR-10 classification gives similar accuracy compared to that of pruning convolutional layer in the main branch alone. Therefore, pruning convolutional layers in the residual branch have little scope for complex classification problems.

Our experiments reveal that the proposed passive filter pruning method reduces computations in computing filter importance significantly and achieves similar accuracy compared to that of the active filter pruning methods without involving dataset.

\section{Conclusion}
\label{sec: conclusion}
We present a passive filter pruning method to reduce computational complexity and memory requirement for convolutional neural networks (CNNs). The proposed pruning method considers the operator norm of the filters to compute filter importance. Various CNNs are pruned to validate  the proposed pruning method and  our experiments reveal that the pruned CNNs give similar accuracy at significant reduction in number of parameters and the number of multiply-accumulate operations compared to that of the unpruned  CNNs. The proposed operator norm based pruning method  generalizes better across CNNs designed for audio and image classification compared to that of entry-wise norm methods. The proposed filter pruning method does not use dataset in computing filter importance, takes less pruning time and achieves similar accuracy compared to that of the feature map based active filter pruning. In addition, the memory footprints due to feature map generation are also not there in the proposed pruning method. 

In future, we would like to make the overall pruning framework computationally efficient by improving the fine-tuning process in recovering the lost performance due to pruning in a faster manner which might use appropriate few examples to fine-tune the pruned network with only few epochs.

\section*{Acknowledgements}
This work was supported by Engineering and Physical Sciences Research Council (EPSRC) Grant EP/T019751/1 \enquote{AI for Sound (AI4S)}. For the purpose of open access, the authors have applied a Creative Commons Attribution (CC BY) licence to any Author Accepted Manuscript version arising. Thanks to Dr. Padmanabhan Rajan (Associate Professor) and Dr. Renu M Rameshan (former Assistant Professor) from IIT Mandi, India for their initial discussions. Thanks to Dr. Yunpeng Li,  Senior Lecturer, University of Surrey, for his suggestions and discussions.

\bibliographystyle{IEEEtran}
\bibliography{ref.bib}

\newpage

\section{Supplementary Material}
\begin{figure}[h]
    \centering
    \includegraphics[scale=0.5]{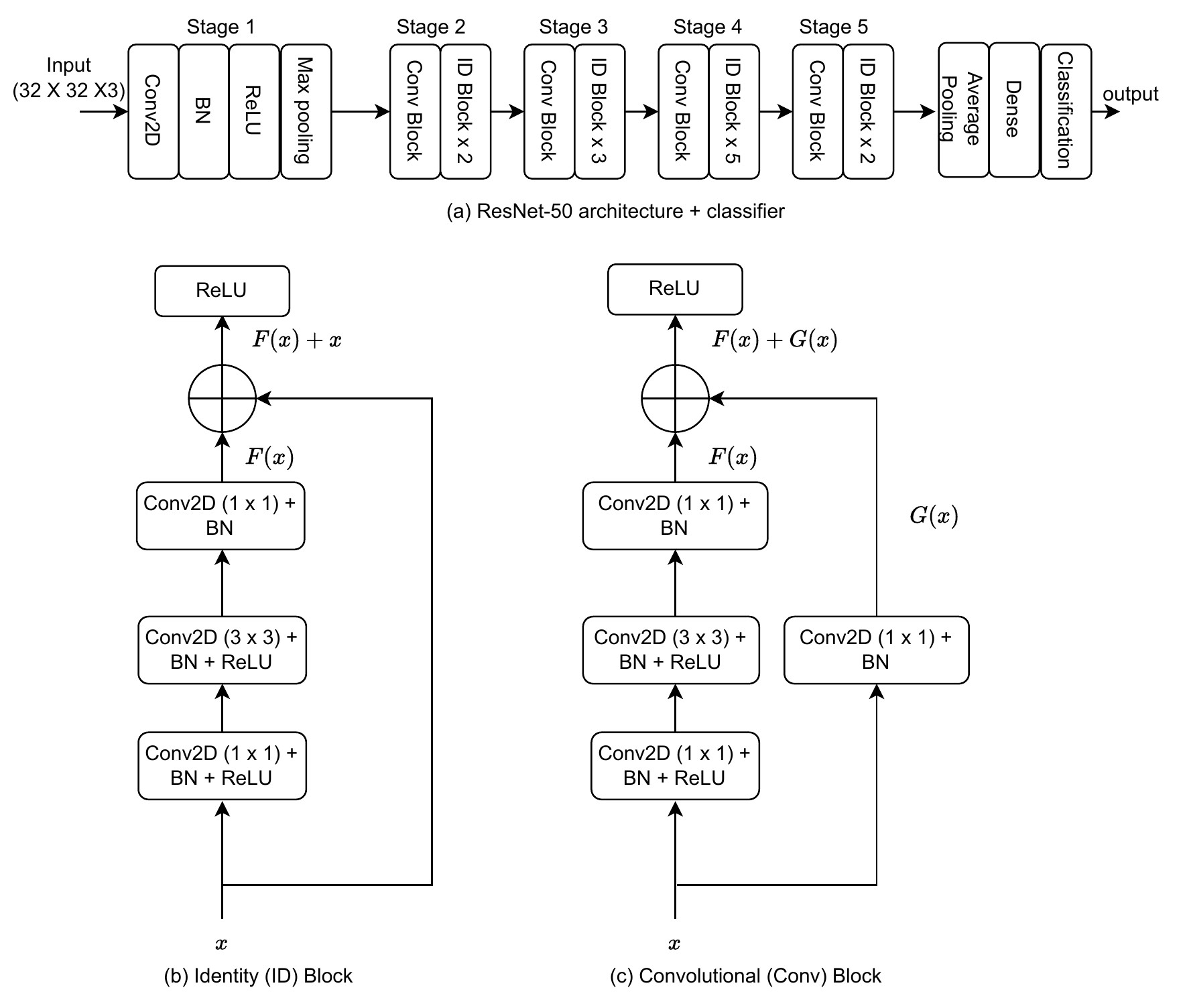}
    \caption{(a) Architecture to classify CIFAR-10 dataset  using ResNet-50 (stage 1 to stage 5) which comprises of (b) identity blocks and (c) convolutional blocks, followed by global average pooling layer, a dense and a classification layer. \enquote{BN} denotes batch normalization. \enquote{ReLU} denotes rectified linear unit activation function and Conv2D is a convolutional layer with 2D filters.}
    \label{fig: resnet architecture for imagenet classifcaiton}
\end{figure}

\begin{figure*}[h]
    \centering\includegraphics[scale=0.4]{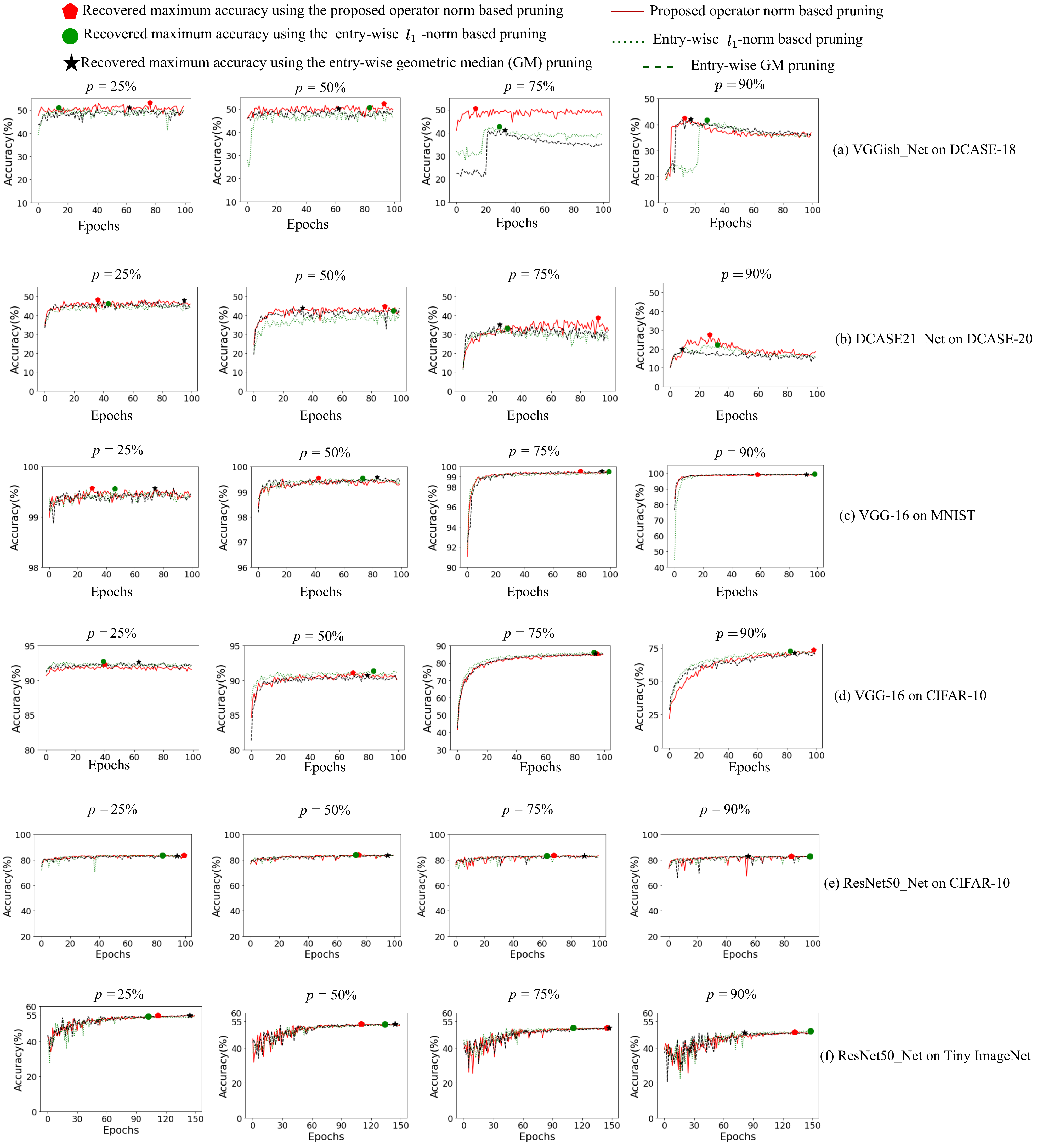}
    \caption{Convergence plots showing regain in the accuracy during fine-tuning of the various pruned  networks obtained at different pruning ratios.}
    \label{fig: convergence plots}
\end{figure*}

In the supplementary material, we provide two figures, Figure \ref{fig: resnet architecture for imagenet classifcaiton} and Figure \ref{fig: convergence plots}.
Figure \ref{fig: resnet architecture for imagenet classifcaiton} shows the architecture of ResNet-50 and Figure \ref{fig: convergence plots} presents the convergence plots for various pruned networks obtained during fine-tuning.

\vfill

\end{document}